\documentclass[10pt,twocolumn,letterpaper]{article}

\usepackage[pagenumbers]{cvpr} 

\usepackage{graphicx}
\usepackage{amsmath}
\usepackage{amssymb}
\usepackage{booktabs}
\usepackage{multirow}
\usepackage{makecell}

\usepackage[pagebackref,breaklinks,colorlinks]{hyperref}

\usepackage[capitalize]{cleveref}
\crefname{section}{Sec.}{Secs.}
\Crefname{section}{Section}{Sections}
\Crefname{table}{Table}{Tables}
\crefname{table}{Tab.}{Tabs.}

\def\cvprPaperID{3597} 
\def\confName{CVPR}
\def\confYear{2022}

\begin{document}
\title{Portrait Eyeglasses and Shadow Removal by Leveraging 3D Synthetic Data}

\author{Junfeng Lyu\quad Zhibo Wang\quad Feng Xu\\
	 School of Software and BNRist, Tsinghua University\\
}
\maketitle

\begin{abstract}
	In portraits, eyeglasses may occlude facial regions and generate cast shadows on faces, which degrades the performance of many techniques like face verification and expression recognition. 
	Portrait eyeglasses removal is critical in handling these problems. However, completely removing the eyeglasses is challenging because the lighting effects (\eg, cast shadows) caused by them are often complex.
	In this paper, we propose a novel framework to remove eyeglasses as well as their cast shadows from face images. 
	The method works in a detect-then-remove manner, in which eyeglasses and cast shadows are both detected and then removed from images. Due to the lack of paired data for supervised training, we present a new synthetic portrait dataset with both intermediate and final supervisions for both the detection and removal tasks. Furthermore, we apply a cross-domain technique to fill the gap between the synthetic and real data.
	To the best of our knowledge, the proposed technique is the first to remove eyeglasses and their cast shadows simultaneously. The code and synthetic dataset are available at \url{https://github.com/StoryMY/take-off-eyeglasses}.
\end{abstract}

\begin{figure}[htp]
	\centering
	\includegraphics[width=1.0\linewidth]{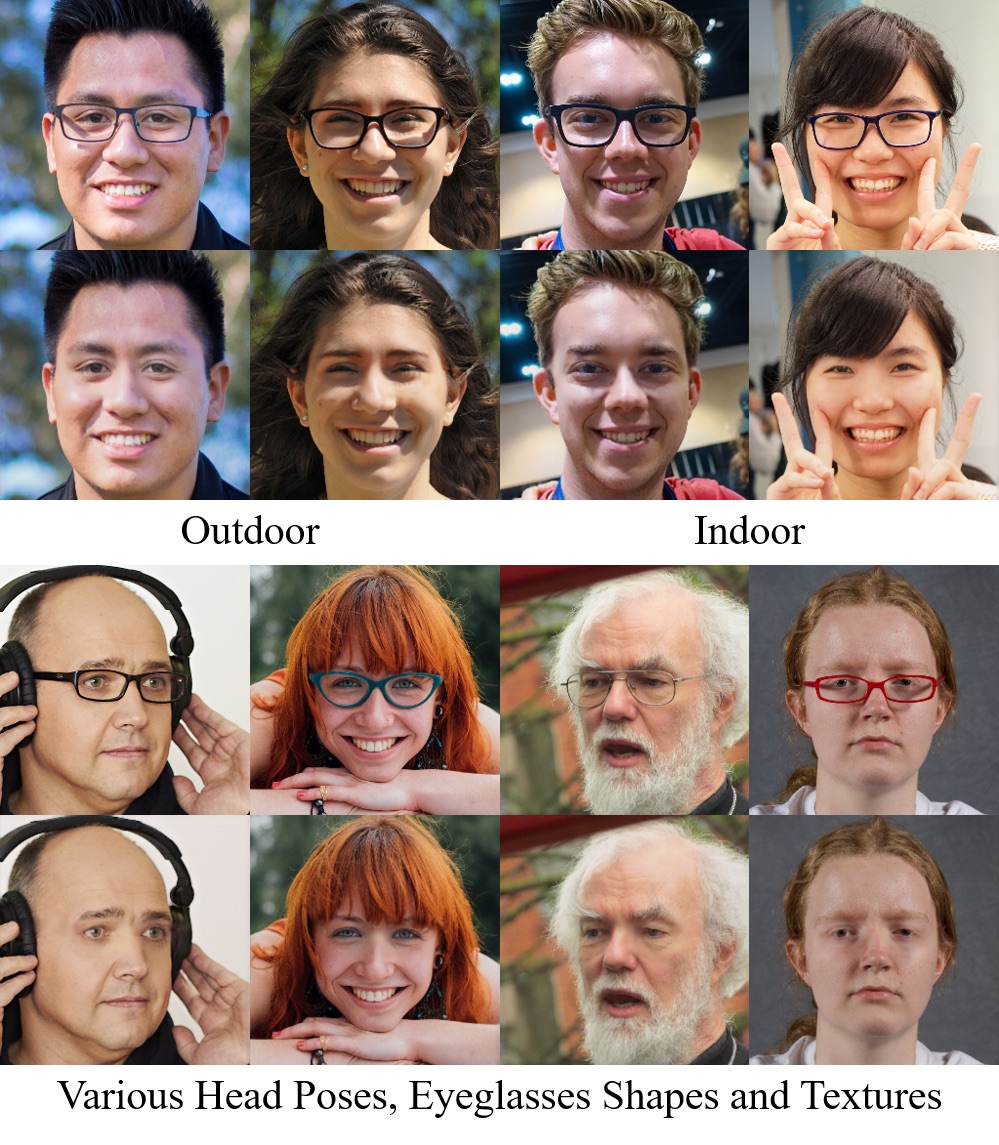}
	\vspace{-2em}
	\caption{Our method allows to remove eyeglasses and their shadows simultaneously. It produces photo-realistic results under various illuminations, head poses and eyeglasses with different shapes and textures.}
	\label{fig:opendoor}
\end{figure}

\section{Introduction}

A large portion of people wears eyeglasses in their daily lives. In their face photos, eyeglasses usually bring unwanted occlusions and cast shadows on faces, which lead to inaccuracy in many useful techniques like image-based face verification\cite{taigman2014deepface,schroff2015facenet}, expression recognition\cite{tang2013deep}, fatigue detection\cite{dwivedi2014drowsy,hajinoroozi2015prediction,9304573}, \etc.
Besides, in photography, removing eyeglasses from portraits could be needed for aesthetic reasons, giving users a choice to edit their portraits. Therefore, it is beneficial to develop an automatic technique for portrait eyeglasses removal.

However, completely removing eyeglasses suffers some key challenges. First, to recover the occluded facial region and keep it consistent with the remaining regions is a difficult task as facial skin has rich details and complex reflectance. Second, only recovering the occluded region cannot ensure visually convincing results as eyeglasses also bring various lighting effects (\eg, cast shadows, reflections and distortions) on face regions. 
Explicitly modeling these effects is extremely difficult as the physical rules to generate these effects are complicated. And it requires a delicate perception of the eyeglasses geometry, face geometry and lighting conditions, which are also difficult to obtain from a single portrait. 

Recently, deep learning\cite{krizhevsky2012imagenet,simonyan2014very} has shown its great potential in handling tasks related to face editing\cite{gao2021high,Li_2021_CVPR}, and has been successfully applied to portrait eyeglasses removal\cite{hu2020unsupervised} with the help of the face datasets\cite{liu2015faceattributes,karras2018progressive} containing eyeglasses labels. However, these techniques only focus on the eyeglasses but not the corresponding lighting effects. 
ByeGlassesGAN\cite{lee2020byeglassesgan} constructs paired data containing some lighting effects for training. However, as it uses 2D methods to synthesize the data, the quality and realism are quite limited. Also, it does not take cast shadows into consideration.

In this paper, we propose a novel eyeglasses removal technique using a synthetic dataset which considers 3D shadows and uses a cross-domain training strategy to fill the gap between synthetic and real data. 
This method jointly removes eyeglasses and their cast shadows, generating more visually plausible results compared to the previous state-of-the-art methods.
In order to facilitate learning the relation between eyeglasses and cast shadows, we introduce a novel mask-guided multi-step network architecture for eyeglasses removal.
The proposed network first detects two masks for both eyeglasses and their cast shadows. Then, the estimated masks are used as guidance in the multi-step eyeglasses removal.
We observe that the shadows to be removed are caused by the eyeglasses, and we use this fact to carefully construct our network where the eyeglasses and shadows are handled in well-designed orders in both the detection and removal tasks. In this way, the network can well take eyeglasses as an important prior when dealing with the shadows.

For training this network, we build a photo-realistic synthetic dataset using high-quality face scans collected by \cite{10.1145/3414685.3417824} and 3D eyeglasses models made by artists, with principled BSDF\cite{mcauley2012practical} to achieve high rendering quality. This dataset contains a large amount of data for supervised training, covering various identities, expressions, eyeglasses, and illuminations. Another benefit of using the synthetic dataset is that we can synthesize images that cannot be captured in real world, i.e., images with eyeglasses but no shadows and images with shadows but no eyeglasses. These images can be used as intermediate supervisions to train the proposed network.

Although the accurate 3D information and the high-end rendering technique improve the photo-realism of our synthetic data, the network still cannot generalize well to real images due to the gap between the synthetic and real domain. Inspired by \cite{huang2017arbitrary} and \cite{tsai2018learning}, we develop a cross-domain segmentation module that leverages a real image dataset to build a uniform domain for both the real and synthetic images. This helps to prevent the proposed network from using domain-specific information to detect eyeglasses and their cast shadows.

In summary, our main contributions are listed as follows:
\begin{itemize}
	\item We design a novel mask-guided multi-step network architecture which is the first attempt in the literature to remove both eyeglasses and their cast shadows from portraits and achieves high realism. 
	\item We present a high-quality synthetic portrait dataset which provides both intermediate and final supervisions for training eyeglasses/shadows detection and removal networks.
	\item We introduce a cross-domain segmentation module to enhance the generalization capability on real face images. 
\end{itemize}

\section{Related Works}

\textbf{Eyeglasses Removal.} Early works\cite{1262319,1407883,yi2011learning,wong2013eyeglasses,1640924} remove eyeglasses by statistical learning. The key assumption of these works is that the facial regions occluded by eyeglasses can be reconstructed from other faces without eyeglasses. However, these methods usually assume frontal faces and controlled environments, which limits their applications. Later works, \eg, ERGAN\cite{hu2020unsupervised} and ByeGlassesGAN\cite{lee2020byeglassesgan}, use deep neural networks in eyeglasses removal.
ERGAN\cite{hu2020unsupervised} proposes an unsupervised architecture for eyeglasses removal in the wild, while ByeGlassesGAN\cite{lee2020byeglassesgan} manually constructs paired data and propose a multi-task framework for eyeglasses detection and removal. These methods can successfully remove eyeglasses in more general application scenarios. However, cast shadows caused by eyeglasses are often ignored in both methods as they do not explore the connections between eyeglasses and cast shadows. Unlike these methods, we found that by developing an architecture to learn this connection, the network can remove the eyeglasses and their cast shadows at the same time, generating more visually convincing results.

\begin{figure*}[htp]
	\centering
	\includegraphics[width=0.98\linewidth]{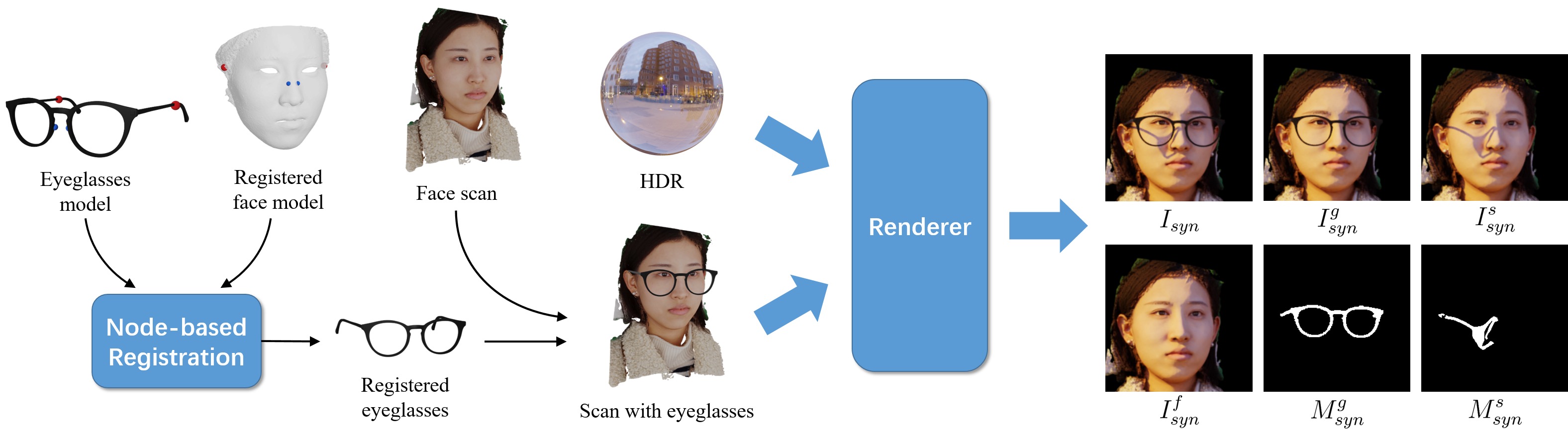}
	\vspace{-1em}
	\caption{Illustration of portrait synthesis. We define two fix nodes (red), two floating nodes (blue) on the registered face model and their corresponding vertices on each eyeglasses model. With node-based registration, we compute a plausible pose to align the eyeglasses model with the face scan. Then, we combine them with a HDR lighting to render our synthetic data: $I_{syn}$, $I_{syn}^{g}$, $I_{syn}^{s}$, $I_{syn}^{f}$, $M_{syn}^{g}$ and $M_{syn}^{s}$.}
	\label{fig:synthesis}
	\vspace{-1.em}
\end{figure*}

\textbf{Face Attributes Manipulation.} Facial image manipulation techniques \cite{shen2017learning,bao2018towards,gao2021high} have been developed rapidly in recent years. Most of them jointly solve multi-label\cite{choi2018stargan,8718508,liu2019stgan,Wu2019RelGANMI,choi2020starganv2} or multi-style\cite{zhu2017multimodal,huang2018munit,lee2018diverse,almahairi2018augmented,guo2019mulgan} issues. DFI\cite{upchurch2017deep} manipulates face attributes via interpolation of different feature vectors. AttGAN\cite{8718508} manipulates facial images via attribute classification constraint and reconstruction learning. STGAN\cite{liu2019stgan} incorporates difference attribute vector and selective transfer units (STUs) for arbitrary attribute editing. HiSD\cite{Li_2021_CVPR} proposes a hierarchical style disentanglement framework for image-to-image translation, which organizes the labels using a hierarchical tree structure and overcomes the disadvantages of previous joint methods\cite{romero2019smit,wang2019sdit,yu2019multi,almahairi2018augmented,xiao2017dna,zhou2017genegan,Xiao_2018_ECCV}. Additionally, some works\cite{tewari2020stylerig,deng2020disentangled} combine 3D Morphable Model (3DMM) with StyleGAN\cite{karras2019style} to control facial images semantically. We found that manipulating external attributes (\eg, hats or eyeglasses) is more difficult than manipulating internal attributes of faces as facial accessories often lead to occlusions or extra lighting effects (\eg, cast shadows).
Unlike the previous works, we focus on eyeglasses removal and aim to remove not only the eyeglasses but also their corresponding cast shadows.

\textbf{Domain Adaptation for Segmentation.} The majority of works in this task are usually designed for urban scenes. \cite{hoffman2016fcns} combines both global and local alignment with a domain adversarial training. \cite{zhang2017curriculum} uses curriculum learning to address the domain adaptation. \cite{chen2017no} proposes an unsupervised method to adapt segmenters across different cities. Other works \cite{tsai2018learning,chen2019learning} apply discriminators on the output space to align source and target segmentation, while \cite{zhu2018penalizing} utilizes a conservative loss to naturally seek the domain-invariant representations. FDA\cite{yang2020fda} proposes a novel method that solves the domain adaptation via a simple Fourier Transform and its inverse. Based on the aforementioned methods, we additionally consider the relevance between the eyeglasses and cast shadows, and successfully bridge the gap between synthetic and real face images.

\section{Portrait Synthesis with Eyeglasses}
\label{sec:data_synthesis}

In order to build paired data for supervised training, we use 3D rendering to generate synthetic images. As shown in \cref{fig:synthesis}, we first make the face scan ``wear'' the 3D eyeglasses via node-based registration. Then, we render the scan with eyeglasses under a randomly chosen illumination. By setting the eyeglasses or their cast shadows to be visible or invisible, we can get four different types of rendered images. The masks of the eyeglasses and the cast shadows are also generated. Details are described as follows.

\subsection{Data Preparation} 

For 3D face data, we directly use the dataset collected by \cite{10.1145/3414685.3417824}. This dataset contains the face scans of 438 subjects with 20 expressions, varying from male to female and young to old. In addition to raw scans, we also acquire the registered face models with the same topology. For 3D eyeglasses models, we ask professional artists to create 21 eyeglasses models, which contain various shapes and textures.

\subsection{Eyeglasses Alignment}

In order to put eyeglasses on the plausible positions of the face, we manually label four anchor nodes ($\boldsymbol{A}_i, i\in\{1,2,3,4\}$) on each eyeglasses model and their corresponding vertices ($\boldsymbol{V}_i, i\in\{1,2,3,4\}$) on the template face model used for registration. Specifically, these four nodes consist of two fixed nodes on the face temples and two floating nodes on both sides of the nose as shown in \cref{fig:synthesis}.
Then, we compute the rotation $\boldsymbol{R}\in SO(3)$, the translation $\boldsymbol{t}\in \mathbb{R}^3$ and the scaling $s\in \mathbb{R}$ by minimizing the distance between the nodes and their corresponding vertices using Singular Value Decomposition \cite{quan1999linear}, expressed as
\begin{equation}
	E(\boldsymbol{R}, \boldsymbol{t}, s; \boldsymbol{A}_i, \boldsymbol{V}_i)=\sum_{i=1}^{4}\|s\cdot \boldsymbol{R}\boldsymbol{A}_i+\boldsymbol{t}-\boldsymbol{V}_i\|^2_2.
\end{equation}

According to our observation, people put their eyeglasses on different nose positions. To enrich the wearing styles of our synthetic data, we define various candidate pairs of floating nodes in the nose region of the face template and randomly choose one pair for eyeglasses alignment. Also, we randomly change the color of eyeglasses to enrich their textures.

\begin{figure*}[tbp]
	\centering
	\includegraphics[width=1.0\linewidth]{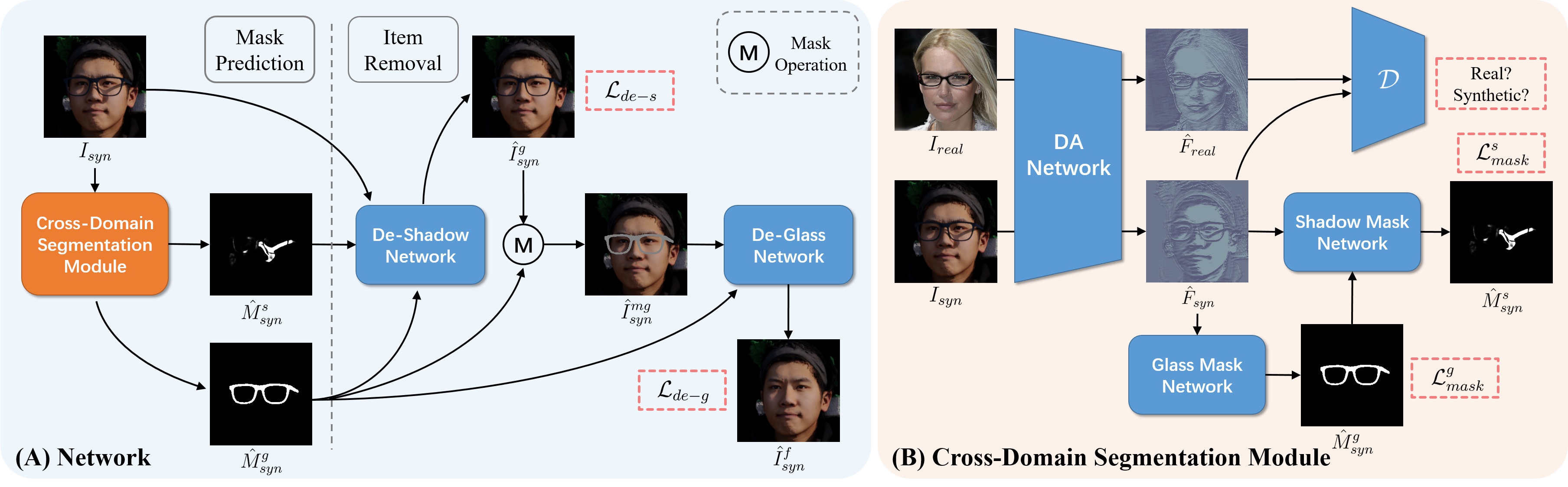}
	\vspace{-2.3em}
	\caption{Illustration of the proposed network architecture. 
		\textbf{(A)} Our network includes two stages: mask prediction stage and item removal stage. The mask prediction stage aims to estimate the eyeglasses mask and shadow mask via a cross-domain segmentation module. In the item removal stage, we successively employ a De-Shadow Network and a De-Glass Network to remove cast shadows and eyeglasses with the guidance of the two predicted masks.
		\textbf{(B)} In the cross-domain segmentation module, the Domain Adaptation (DA) Network normalizes the input images to uniform feature maps with the help of a discriminator. Then, the Glass Mask Network and Shadow Mask Network take the uniform feature maps to predict eyeglasses and shadows masks, respectively.
	}
	\label{fig:pipeline}
\end{figure*}

\subsection{Rendering Setting}

The data variety and photo-realism are well considered in the portrait rendering. In detail, we first collect 367 HDR lightings from \textit{Poly Heaven}\footnote{\href{https://polyhaven.com/}{https://polyhaven.com/}\label{url:pr_dataset}} to increase the diversity of illuminations. During the rendering, the lighting variation is further augmented by setting a random rotation of the global scene. Besides, we render each face scan which randomly ``wears'' a pair of eyeglasses by a random head pose.
For photo-realistic synthesis, we use the principled BSDF implementation in Blender to render our synthetic data with the rendering setting empirically adjusted by a professional artist.

For each rendering sample, we render four kinds of images with different visibility combinations of the eyeglasses and their cast shadows: $I_{syn}$, $I_{syn}^{g}$, $I_{syn}^{s}$, $I_{syn}^{f}$. The eyeglasses mask $M_{syn}^{g}$ and the shadow mask $M_{syn}^{s}$ are also synthesized in the rendering.

\section{Portrait Eyeglasses Removal Network}

The architecture of the proposed network is illustrated in \cref{fig:pipeline}. Our network is designed based on the following considerations: 
\textit{1)} ByeGlassesGAN\cite{lee2020byeglassesgan} improves eyeglasses removal with a parallel segmentation task, which has demonstrated the importance of mask prediction in eyeglasses removal. 
Inspired by their method, we remove the eyeglasses in a more natural way by first explicitly detecting the eyeglasses in a \textit{mask prediction stage} and then removing eyeglasses with the guidance of predicted masks in an \textit{item removal stage}.
\textit{2)} We further enhance the eyeglasses removal performance by using a multi-step strategy in both above stages to treat the eyeglasses and their cast shadows in sequential order. Considering that the shadows to be removed are caused by eyeglasses, the eyeglasses should be a guidance in both shadow mask prediction and shadow removal. 
\textit{3)} The proposed network is trained to remove eyeglasses using the synthetic dataset. To make it generalized to real images, we use a Domain Adaptation (DA) Network to convert the input images into uniform feature maps. The uniform feature maps eliminate the domain-specific information to confuse a discriminator but retain the structural information for eyeglasses and shadow mask prediction.

\subsection{Mask Prediction Stage}
Given an input portrait $I$ with eyeglasses, our method estimates the  eyeglasses mask $\hat{M}^g$ and shadows mask $\hat{M}^s$ in the mask prediction stage using a cross-domain segmentation module. This module is composed of a DA Network, a Glass Mask Network and a Shadow Mask Network. 

In order to tackle the gap between synthetic and real domain, the DA Network is trained to transfer the input image $I$ to a uniform domain, outputting the uniform feature map $\hat{F}$. Inspired by \cite{huang2017arbitrary} and \cite{tsai2018learning}, we apply adversarial learning to find the uniform domain assisted with a discriminator $\mathcal{D}$. This discriminator $\mathcal{D}$ is trained to distinguish whether the feature map $\hat{F}$ is from a real image or a synthetic image while the DA Network aims to fool the discriminator. We utilize 
LSGAN\cite{mao2017least,CycleGAN2017} for more stable training:
\begin{equation}
	\mathcal{L}^\mathcal{D}_{adv}=(\mathcal{D}(\hat{F}_{syn}))^2+(\mathcal{D}(\hat{F}_{real})-1)^2,
\end{equation}
\begin{equation}
	\mathcal{L}^\mathcal{G}_{adv}=(\mathcal{D}(\hat{F}_{syn})-1)^2,
\end{equation}
where $\hat{F}_{real}$ and $\hat{F}_{syn}$ are the corresponding feature maps of real and synthetic data. Specifically, the DA Network consists of the first layer of  
a pre-trained VGG encoder\cite{vgg} with fixed parameters, combined with six trainable ResNet blocks\cite{he2016deep} with instance normalization.

We use a multi-step strategy to predict the eyeglasses mask $\hat{M}^g$ and the corresponding shadow mask $\hat{M}^s$ from the uniform domain feature $\hat{F}$. Instead of extracting these two masks together using a single network, we first estimate the eyeglasses mask $\hat{M}^g$ using a Glass Mask Network. Then, the previous outputs $\hat{F}$ and $\hat{M}^g$ are together fed into a Shadow Mask Network to predict the shadow mask $\hat{M}^s$, with the consideration that the eyeglasses masks could be a guidance in the shadow mask prediction.
We learn the eyeglasses mask $M^g$ and the cast shadow mask $M^s$ in a supervised manner as follows,
\begin{eqnarray}
	&\mathcal{L}_{mask}^g = L_{\mathcal{E}}(M_{syn}^{g}, \hat{M}_{syn}^{g}),\\ &\mathcal{L}_{mask}^s = L_{\mathcal{E}}(M_{syn}^{s}, \hat{M}_{syn}^{s}), \\
	&L_{\mathcal{E}}(M, \hat{M}) = -M \log \hat{M} -(1-M)\log (1-\hat{M}),
\end{eqnarray}
where $L_{\mathcal{E}}$ is the widely used binary cross entropy (BCE) loss. Experiments in \cref{sec:abs} demonstrate that with the guidance of the estimated eyeglasses mask $\hat{M}^g$, the predicted shadow mask $\hat{M}^s$ will be more complete.

Overall, the training loss for the mask prediction stage is formulated as
\begin{equation}
	\begin{aligned}
		\mathcal{L}_{predict} &= \lambda_{adv} \mathcal{L}^\mathcal{D}_{adv} + \lambda_{adv}\mathcal{L}^\mathcal{G}_{adv} \\
		&+ \lambda_{mask} \mathcal{L}_{mask}^g + \lambda_{mask} \mathcal{L}_{mask}^s,
	\end{aligned}
	\label{eq:MP}
\end{equation}
where $\lambda_{adv}$ and $\lambda_{mask}$ are the weights for adversarial learning and mask prediction, respectively. 

\subsection{Item Removal Stage}

This stage aims to remove eyeglasses and cast shadows, and we call it \textit{item removal stage} for short. It takes the two predicted masks as clues to achieve clean eyeglasses and shadow removal. When removing these items, we also apply the multi-step strategy. However, different from the multi-step setup used in the mask prediction stage, in which our method first handles eyeglasses and then shadows, we deal with eyeglasses and shadows in an inverse order in this stage. 
This is because if we first remove the eyeglasses, the network will lose the abundant indications of shadow intensity and locations.

With an input image $I$, we first use a De-Shadow Network to remove the cast shadows of the eyeglasses. To help the network better locate the cast shadows to be removed, the estimated eyeglasses mask $\hat{M}^g$ and shadow mask $\hat{M}^s$ are also fed to the De-shadow Network. In order to learn the shadow-removed image $I^g$, we employ a $L_1$ regression loss, written as
\begin{equation}
	\mathcal{L}_{de-s}=\|\hat{I}_{syn}^{g}-I_{syn}^{g}\|_1,
\end{equation}
where $\hat{I}_{syn}^g$ indicates the output of our De-Shadow Network.

After removing the cast shadows, we use a De-Glass Network to further remove the eyeglasses in the next step. The large variety of eyeglasses textures in real world will lower the performance of eyeglasses removal. 
To enhance the robustness of our method, we adopt a mask operation to set the pixel values of the eyeglasses regions to $\boldsymbol{0}$. This operation eliminates the texture of eyeglasses from $\hat{I}^g$, forcing the De-Glass Network to remove the eyeglasses only according to the structure instead of textures.
Finally, the De-Glass Network takes the masked shadow-removed result $\hat{I}^{mg}$ and the estimated eyeglasses mask $\hat{M}^g$ as input and learns the eyeglasses-removed image $I^f$ via the following constraint:
\begin{equation}
	\mathcal{L}_{\mathit{de-g}}=\|\hat{I}_{syn}^{f}-I_{syn}^{f}\|_1.
\end{equation}
where $\hat{I}^f_{syn}$ represents the output of our De-Glass Network.

To sum up, the training loss for the item removal stage is formulated as
\begin{equation}
	\mathcal{L}_{\mathit{remove}}=\lambda_{de-s}\mathcal{L}_{de-s}+\lambda_{de-g}\mathcal{L}_{de-g},
	\label{eq:IR}
\end{equation}
where $\lambda_{de-s}$ and $\lambda_{de-g}$ are the weights for shadow and eyeglasses removal, respectively.

\section{Experiments}

In this section, we first describe the datasets and our implementation details. Then, we compare our method with the state-of-the-art eyeglasses removal and image-to-image translation methods qualitatively and quantitatively. 
Finally, we evaluate the key contributions of the proposed method via ablation study. Note that besides the results in \cref{fig:opendoor}, we will show more various results in our supplementary material.

\textbf{Dataset.} We use our synthetic dataset described in \cref{sec:data_synthesis} and CelebA\cite{liu2015faceattributes} to train the proposed network. 
For synthetic dataset, we randomly sample 73 identities of the 438 identities. Each identity contains 20 face scans with different expressions. We combine the face scans randomly with 5 eyeglasses and 4 HDR lightings, finally generating 29,200 training samples.
CelebA is a real-world portrait dataset that contains 202,599 face images of 10,177 identities and is annotated with 5 landmarks and 40 binary attributes for each image. Using the attributes labels, we split 13,193 images with eyeglasses and 189,406 images without eyeglasses from it. Additionally, we adopt FFHQ\cite{karras2019style} and MeGlass\cite{guo2018face} for testing. FFHQ contains 70,000 high-quality portraits and it also covers accessories like eyeglasses. Using face parsing\cite{FP}, we roughly split 11,778 images with eyeglasses from it. MeGlass is a dataset containing 1,710 identities and each identity has images with and without eyeglasses. This dataset is essential for identity preservation validation in \cref{subsec:quantitative_res}. We refer to \cite{karras2019style} to align all the images to a size of $256\times256$ using facial landmarks. 

\textbf{Implementation Details.} Our method is implemented with PyTorch. We use Adam optimizer\cite{kingma2014adam} with $\beta_1=0.5$ and $\beta_2=0.999$. The learning rate is 0.0001 and the batch size is 8. For the weights in the objective functions in \cref{eq:MP} and \cref{eq:IR}, we set $\lambda_{adv}=0.1$, $\lambda_{mask}=1$, $\lambda_{de-s}=1$, and $\lambda_{de-g}=1$. Apart from the DA Network and the discriminator, all the other networks utilize the architecture in \cite{Johnson2016Perceptual}. In practice, we first train the cross-domain segmentation module for 30 epochs and fix it when training networks in the item removal stage, which needs 80 epochs. The total training process costs about two days on a single GTX 1080 GPU.  

\begin{figure*}[htp]
	\centering
	\includegraphics[width=0.98\linewidth]{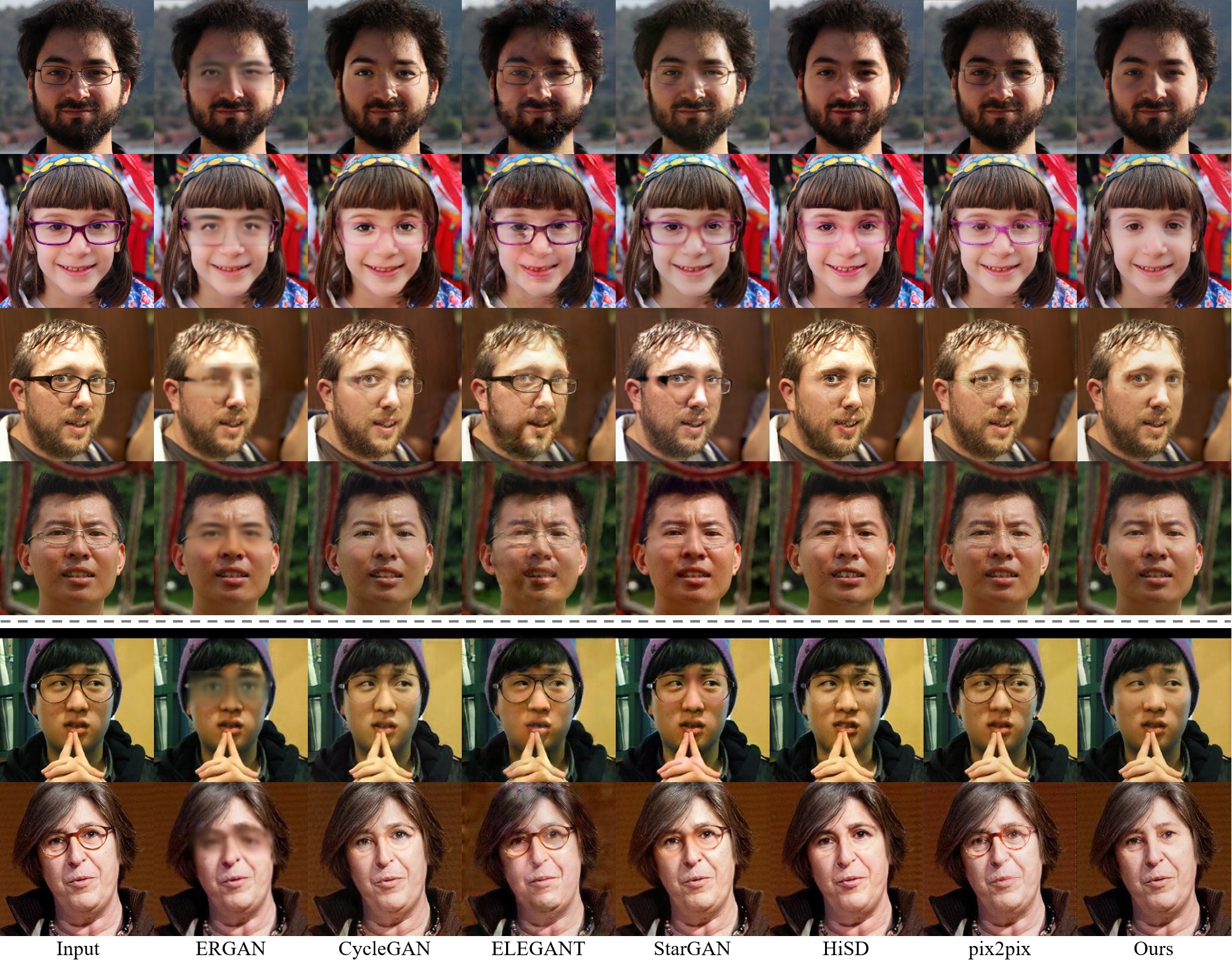}
	\vspace{-1.2em}
	\caption{Qualitative cross-dataset results of different methods on FFHQ dataset (top) and MeGlass (bottom). }
	\label{fig:quality_cmp}
\end{figure*}

\subsection{Comparison with State-of-the-art Methods}
We compare our method with state-of-the-art eyeglasses removal methods: ERGAN\cite{hu2020unsupervised} and ByeGlassesGAN\cite{lee2020byeglassesgan}, as well as image-to-image translation methods including CycleGAN\cite{CycleGAN2017},  StarGAN\cite{choi2018stargan}, ELEGANT\cite{Xiao_2018_ECCV}, pix2pix\cite{isola2017image} and HiSD\cite{Li_2021_CVPR}. 
To ensure fair comparisons, all these methods and our method are not trained on the testing dataset. Specifically, to compare with ERGAN and HiSD, we directly use their released models which are trained on CelebA and CelebA-HQ\cite{karras2018progressive}, respectively. For CycleGAN, StarGAN and ELEGANT, we train them on the task of eyeglasses removal using their codes and the CelebA dataset.
As pix2pix needs paired data, we train it on our synthetic data using the released code. As we cannot reach the authors of ByeGlassesGAN\cite{lee2020byeglassesgan} to conduct a comparison experiment, we just show qualitative comparison using the images posted in their paper. Note that the purpose of the comparisons is not to purely compare different methods in the same setting but to demonstrate which solution better solves the problem.

\vspace{-1em}
\subsubsection{Qualitative Comparison}

\begin{figure}[tbp]
	\centering
	\includegraphics[width=0.92\linewidth]{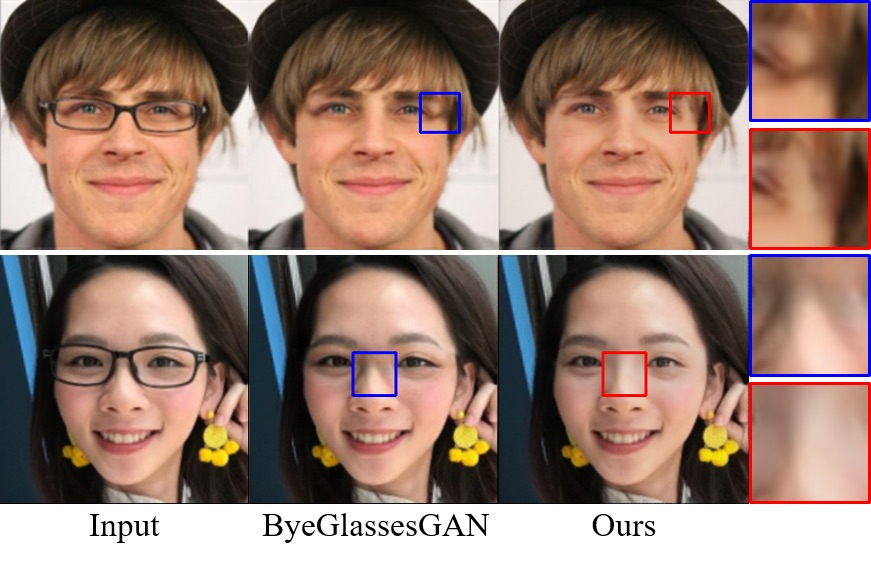}
	\vspace{-0.8em}
	\caption{Qualitative comparison with ByeGlassesGAN\cite{lee2020byeglassesgan} using images from their paper. }
	\label{fig:quality_cmp_bgg}
	\vspace{-1.5em}
\end{figure}

We first compare the visual quality of our method with prior works on various images from FFHQ and MeGlass, covering different ages, genders, head poses, illuminations, eyeglasses shapes and textures. 
As shown in \cref{fig:quality_cmp}, our method achieves the best quality compared to the previous works. 
ELEGANT fails to remove the frames of the eyeglasses on all the test images. ERGAN can remove eyeglasses, but the eyeglasses regions are always blurred. CycleGAN, StarGAN and pix2pix preserve the high-frequency details in the whole eyeglasses regions, but they cannot completely remove eyeglasses for some samples. HiSD seems competitive to ours on some easy samples, but it fails to remove sharp cast shadows (1st row) as well as  
eyeglasses with unusual shape (5th row) and texture (2nd row). Benefiting from the mask-guided learning and our synthetic data, our method can remove various eyeglasses and the corresponding cast shadows. In addition, it generates photo-realistic contents in the regions occluded by eyeglasses or shadows, and retains the consistency with the global illumination and the skin texture of the surrounding regions. 
For ByeGlassesGAN\cite{lee2020byeglassesgan}, we only perform the comparison using the images posted in their paper. Results are shown in \cref{fig:quality_cmp_bgg} and we can see that our method outperforms theirs in the shadow removal.

\subsubsection{Quantitative Results}
\label{subsec:quantitative_res}
For quantitative comparisons, we first use Fr\'{e}chet Inception Distance (FID)\cite{heusel2017gans} to evaluate the realism of the generated images. Then, we apply face recognition technique to evaluate the ability of identity preservation. Finally, we adopt a user study to further evaluate the visual quality of eyeglasses removal.

\textbf{Realism.} 
First, we process the images with eyegalsses in FFHQ by different methods. Then, we compute the FID between the eyeglasses-removed results and the images without eyeglasses in FFHQ. The results (\cref{tab:all_metric}, 1st col) show that our method is competitive with HiSD and outperforms the other methods. This indicates that images generated by ours and HiSD are probably close to the real images without eyeglasses. Note that realism is a subjective measurement that can not be fully represented by FID. For further evaluation, we adopt a \textbf{user study} later. 

\textbf{Identity Preservation.} 
To evaluate the identity preservation ability, we use some metrics commonly used in face recognition\cite{best2014unconstrained,8296998}, including the True Accept Rate at False Accept Rate (TAR@FAR) and Rank-1. To compute these metrics, we first collect 1,227 image triplets from MeGlass dataset. Each triplet contains three images of the same identity: two without eyeglasses and one with eyeglasses. Then, we input the image with eyeglasses into different methods to acquire corresponding eyeglasses-removed results. Finally, we select the first eyeglasses-free image in the triplet as the gallery and all the other images as probes to compute the metrics based on a pre-trained face recognition network\cite{FR}. As shown in \cref{tab:all_metric}, the second eyeglasses-free image in the triplet (\textit{noglass}) achieves high face recognition accuracy as it is a real image containing full identity information. However, the accuracy will degrade when taking the images with eyeglasses as the probe (\textit{glass}), indicating the negative effects of eyeglasses in face recognition. ERGAN, CycleGAN, ELEGANT and pix2pix lead to the further degradation of face recognition after eyeglasses removal while StarGAN and HiSD enhance the metrics. Our method exhibits the most significant increase, which stands for the best ability of eyeglasses removal and identity preservation.

\textbf{User Study.} A user study is conducted to further evaluate the visual quality of eyeglasses removal. In detail, we combine the results of different methods together with the input image to construct a ``question''. Participants are asked to give their opinions based on the visual quality, scoring different results from 1 to 5 (1 for the worst, 5 for the best). In total, we invite 40 participants and each participant is asked to answer 20 randomly sampled ``questions''. As shown in \cref{tab:all_metric}, our method has the highest Mean Opinion Score (MOS), indicating the superiority of our technique.

\begin{table}[tbp]
	\centering
	\footnotesize
	\begin{tabular}{c|ccccc}
		\toprule
		\multirow{2}{*}{} & \multirow{2}{*}{FID$\downarrow$} & \multirow{2}{*}{MOS$\uparrow$}& \multicolumn{2}{c}{TAR@FAR$\uparrow$} & \multirow{2}{*}{Rank-1$\uparrow$} \\
		& & & $1e^{-2}$  & $1e^{-3}$  &   \\
		\midrule
		\textit{glass} & -  & - & 0.6025  & 0.3349  & 0.3716  \\
		\midrule
		ERGAN\cite{hu2020unsupervised}& 38.61  & 1.10 & 0.2839  & 0.1005  & 0.1439  \\
		CycleGAN\cite{CycleGAN2017}& 38.10  & 2.21 & 0.5856  & 0.3186  & 0.3431  \\
		ELEGANT\cite{Xiao_2018_ECCV} & 43.13   & 1.12 & 0.3531   & 0.1507  & 0.1862  \\
		StarGAN\cite{choi2018stargan} & 40.93   & 1.51 & 0.6435   & 0.3773  & 0.4107 \\
		HiSD\cite{Li_2021_CVPR} & \textbf{26.74}   & 3.17 & 0.6329   & 0.3757  & 0.3903  \\
		pix2pix\cite{isola2017image}& 41.42   & 1.52 & 0.5687   & 0.3015  & 0.3422 \\
		Ours & 26.89 & \textbf{4.43} & \textbf{0.6702} & \textbf{0.4315} & \textbf{0.4621}  \\
		\midrule
		\textit{noglass}& - & - & 0.8295   & 0.6430  & 0.6625  \\
		\bottomrule
	\end{tabular}
	\vspace{-0.5em}
	\caption{Quantitative results of different methods.}
	\label{tab:all_metric}
	\vspace{-1.5em}
\end{table}

\subsection{Ablation Study}
\label{sec:abs}
In this subsection, we evaluate the performance of our key contributions in the mask prediction stage and the item removal stage.

\begin{figure}[tbp]
	\centering
	\includegraphics[width=0.98\linewidth]{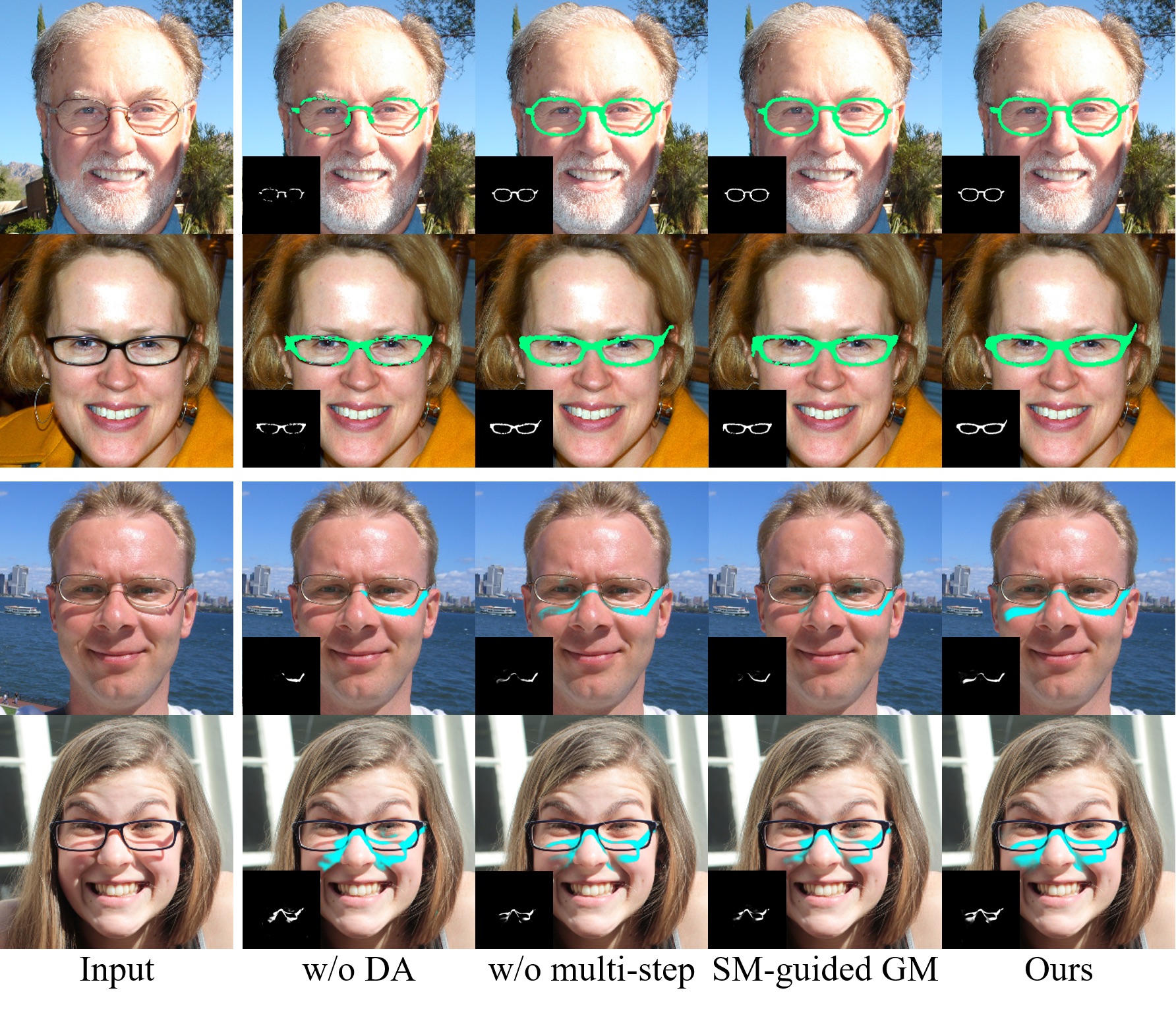}
	\vspace{-1.3em}
	\caption{Visualization of eyeglasses masks (green) and shadow masks (blue) of different ablations in the mask prediction stage. }
	\label{fig:ablation2}
	\vspace{-1em}
\end{figure}

\textbf{Mask Prediction.} 
We first conduct ablation studies for the mask prediction stage.
The first ablation removes the DA Network with two new segmentation networks trained on synthetic data only and tested on real data directly (\textit{w/o DA}). Results in \cref{fig:ablation2} show that without domain adaptation, the estimated eyeglasses masks are sometimes incomplete and thus lead the shadow mask prediction to produce even worse results.
The second ablation removes the multi-step strategy (in mask prediction) by using a single network to estimate the masks of eyeglasses and shadows together (\textit{w/o multi-step}). With the help of the DA network, the eyeglasses masks are properly estimated. However, as the eyeglasses masks can not help the shadow masks estimation in the single-step setting, the estimated shadow masks still have noticeable artifacts.
To further evaluate our assumption that the eyeglasses mask can guide the task of shadow mask prediction as shadows are caused by eyeglasses, we further conduct another ablation setting where the shadow mask is first predicted and then used as guidance in the eyeglasses mask prediction (\textit{SM-guided GM}). Its results 
show that this multi-step setup will lead to worse shadow mask estimation. This further indicates the correctness of our assumption, and the order of the two tasks is important due to the causality between eyeglasses and shadows.

\begin{table}[tbp]
	\centering
	\footnotesize
	\begin{tabular}{c|cccc}
		\toprule
		\multirow{2}{*}{} & \multirow{2}{*}{FID$\downarrow$} & \multicolumn{2}{c}{TAR@FAR$\uparrow$} & \multirow{2}{*}{Rank-1$\uparrow$} \\
		&                      & $1e^{-2}$        & $1e^{-3}$        &                         \\
		\midrule
		w/o DA            & 27.45                & 0.6463       & 0.3977       & 0.4392                  \\
		w/o multi-step    & 27.18                & 0.6683       & 0.4262       & 0.4458                  \\
		SM-guided GM    & 27.30                & 0.6641       & 0.4201       & 0.4523                  \\
		\makecell[c]{GM-guided SM (ours)}           & \textbf{26.89}                & \textbf{0.6702}       & \textbf{0.4315}       & \textbf{0.4621}                 \\
		\midrule
		w/o SM            & 33.89                & 0.6586       & 0.3989       & 0.4327                  \\
		w/o GM            & 42.80                & 0.6567       & 0.3846       & 0.4221                  \\
		w/o multi-step    & 28.66                & 0.6675       & 0.4197       & 0.4498                  \\
		De-Glass First    & 29.58                & 0.6590       & 0.4115       & 0.4417                  \\
		\makecell[c]{De-Shadow First (ours)}          & \textbf{26.89}                & \textbf{0.6702}       & \textbf{0.4315}       & \textbf{0.4621}                 \\
		\bottomrule
	\end{tabular}
	\vspace{-0.8em}
	\caption{Quantitative comparison of different ablations in the mask prediction stage (top) and item removal stage (bottom).}
	\label{tab:ablation_metrics}
	\vspace{-1.7em}
\end{table}

\begin{figure}[tbp]
	\centering
	\includegraphics[width=0.98\linewidth]{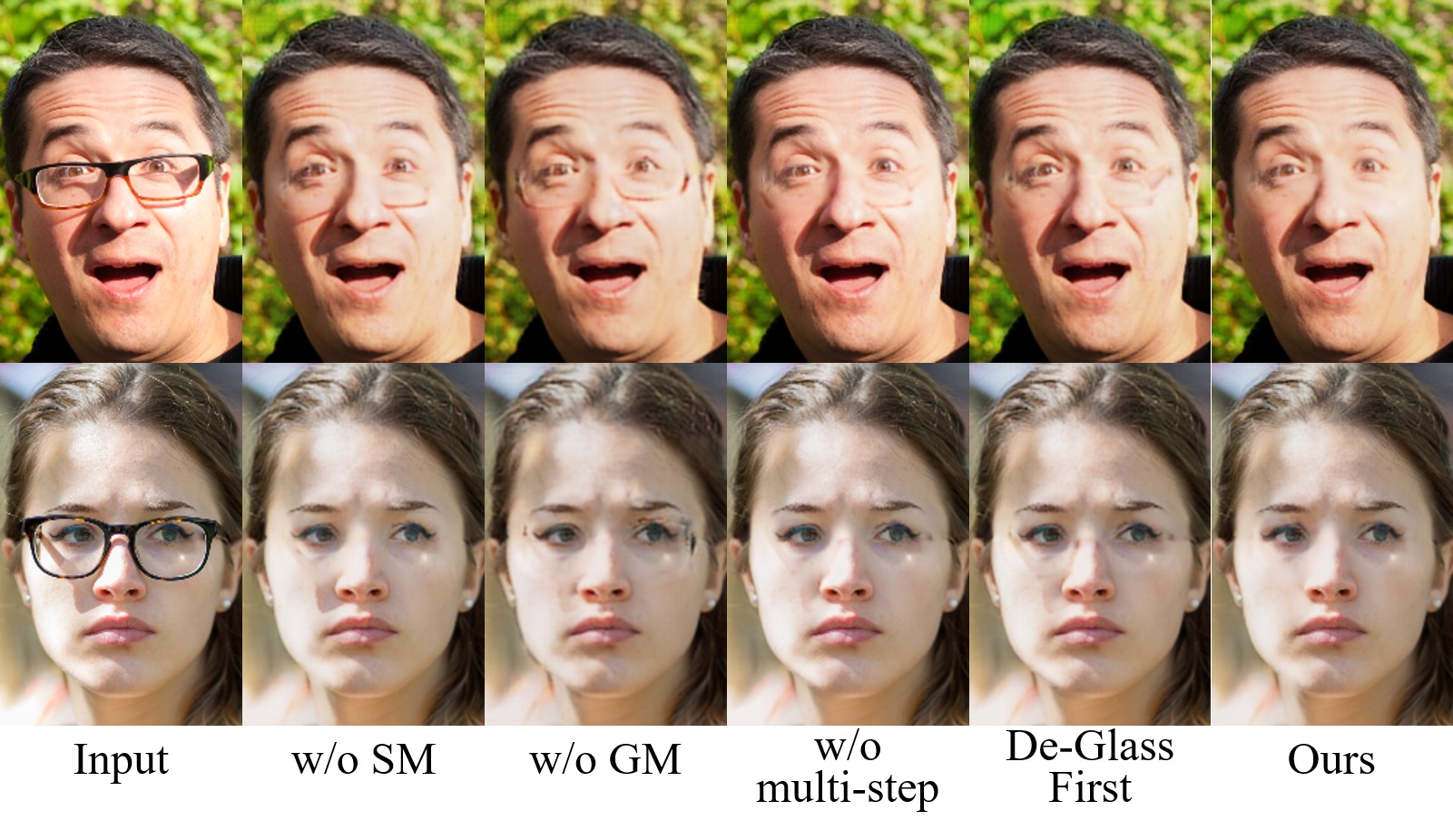}
	\vspace{-1.5em}
	\caption{Qualitative results of different ablations in the item removal stage. }
	\label{fig:ablation1}
	\vspace{-1em}
\end{figure}

\textbf{Item Removal.} Here, we evaluate the effect of mask guidance and the multi-step strategy (in item removal) by comparing different ablation settings. 
We first train two ablation settings without using the shadow mask or the eyeglasses mask (\textit{w/o SM} and \textit{w/o GM}), respectively. 
We also remove eyeglasses and shadows using one network to construct the third ablation setting (\textit{w/o multi-step}). 
Similar to the mask prediction stage, we also invert the order of De-Shadow and De-Glass Network to get the fourth setting (\textit{De-Glass First}). 
Qualitative results in \cref{fig:ablation1} obviously show that \textit{w/o SM} is weak at shadow removal while \textit{w/o GM} fails to remove the complete eyeglasses. Besides, \textit{w/o multi-step} and \textit{De-Glass First} also have noticeable degradation compared to the proposed method. Quantitative results in \cref{tab:ablation_metrics} also manifest the advantages of the proposed method.

\section{Limitations}

\begin{figure}[tbp]
	\centering
	\includegraphics[width=0.98\linewidth]{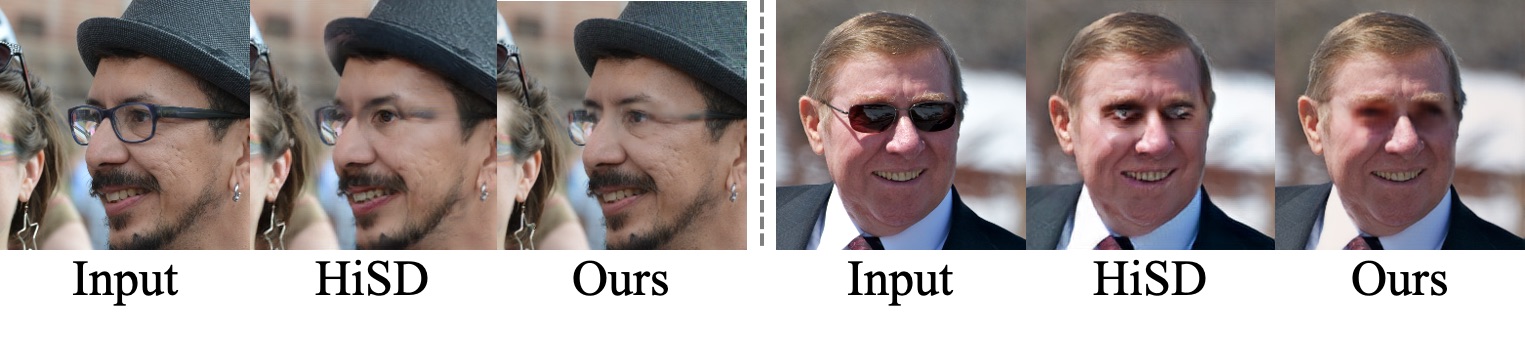}
	\vspace{-1.5em}
	\caption{Limitations. Extreme head pose with effects of lenses (left) and colored lenses (right). These cases are difficult for most of the existing methods. Here, we only show comparisons to the most competitive method (HiSD). }
	\label{fig:limitation}
	\vspace{-1.5em}
\end{figure}

Extensive experiments have shown that the proposed method achieves promising performance on real-world images across age, gender, head pose, illumination and eyeglasses. However, it currently does not perform well on images with extreme head pose or eyeglasses with colored lenses as shown in \cref{fig:limitation}. A large head pose often results in extreme lens distortion, which is expensive to simulate in the portrait synthesis.
Eyeglasses with colored lenses, \eg sunglasses, are still difficult to remove due to the complete occlusions of eyes.
A possible solution is to add more samples of these cases into the training dataset, which will be included in our future work.

\section{Conclusion}

In this paper, we introduce a novel eyeglasses removal technique that first detects and then removes the eyeglasses using the mask-guided multi-step network architecture. To our best knowledge, the proposed method is the first attempt to remove the eyeglasses and their cast shadows simultaneously from a single portrait. Besides, we build a high-quality synthetic portrait dataset, which provides intermediate and final supervisions. In order to fill the gap between the synthetic and real domain, we apply the cross-domain segmentation module to predict the masks of eyeglasses and their cast shadows from a uniform domain for removal guidance. Both qualitative and quantitative experiments demonstrate that our method better preserves the original identity and achieves high realism on real portraits.

\noindent
\footnotesize{\textbf{Acknowledgements.} We thank SenseTime Group Limited for providing computing resources. This work was supported by Beijing Natural Science Foundation (JQ19015), the NSFC (No.61727808, 62021002), the National Key R\&D Program of China 2018YFA0704000. This work was supported by the THUIBCS and BLBCI. Feng Xu is the corresponding author.}

{\small
	\bibliographystyle{ieee_fullname}
	\bibliography{egbib}
}

\clearpage
\appendix
\normalsize

\section{Ablation Study for Mask Operation}
\begin{figure}[btp]
	\centering
	\includegraphics[width=1.0\linewidth]{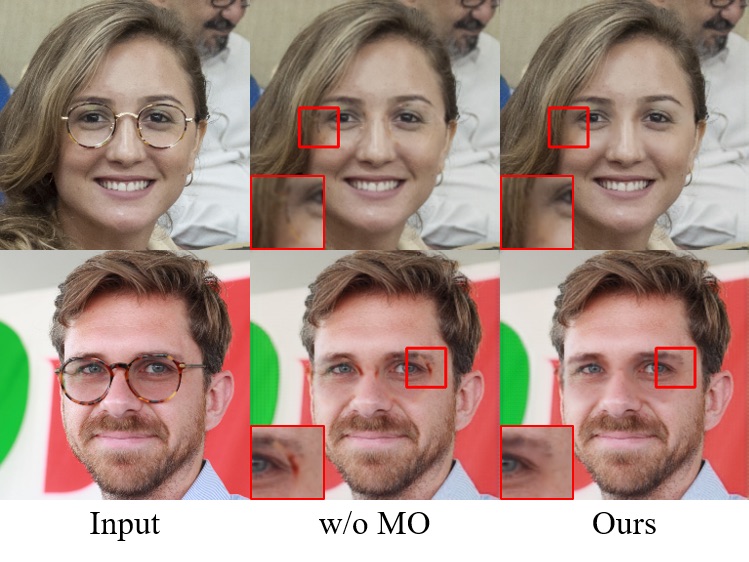}
	\caption{Evaluations on the mask operation.}
	\label{fig:ablation_mo}
\end{figure}

To further evaluate the effect of mask operation in the item removal stage, we additionally train an ablation that directly feeds the shadow-removed results $\hat{I}_{syn}^g$ (\textit{w/o MO}) to the De-Glass Network. 
Qualitative results in \cref{fig:ablation_mo} show that without mask operation, the network has worse results on real-world portraits with unusual eyeglasses textures. 
Quantitative results in \cref{tab:ablation_mo} also show degradation of the realism metric FID and the identity preservation metrics without the mask operation.

\begin{table}[hbp]
	\centering
	\begin{tabular}{c|cccc}
		\toprule
		\multirow{2}{*}{} & \multirow{2}{*}{FID$\downarrow$} & \multicolumn{2}{c}{TAR@FAR$\uparrow$} & \multirow{2}{*}{Rank-1$\uparrow$} \\
		&                      & $1e^{-2}$        & $1e^{-3}$        &                         \\
		\midrule
		w/o MO            & 27.57                & 0.6540       & 0.4095       & 0.4311                  \\
		w/ MO (ours) & \textbf{26.89}                & \textbf{0.6702}       & \textbf{0.4315}       & \textbf{0.4621}                 \\
		\bottomrule
	\end{tabular}
	\vspace{-0.8em}
	\caption{Quantitative comparison.}
	\label{tab:ablation_mo}
\end{table}

\section{More Results}
\begin{figure*}[htp]
	\centering
	\includegraphics[width=1.0\linewidth]{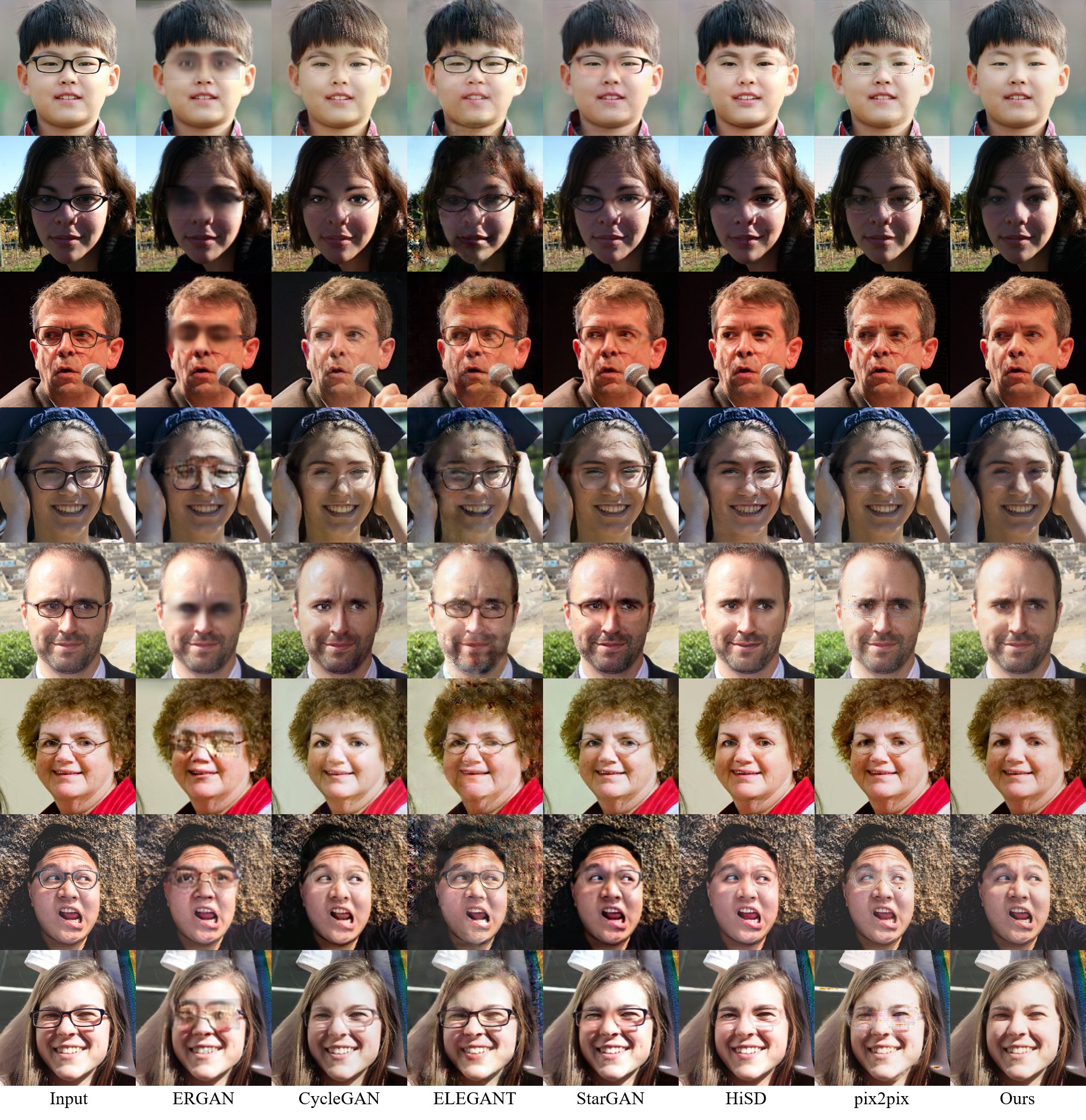}
	\caption{More results of different illuminations.}
	\label{fig:shadows}
\end{figure*}

\begin{figure*}[htp]
	\centering
	\includegraphics[width=1.0\linewidth]{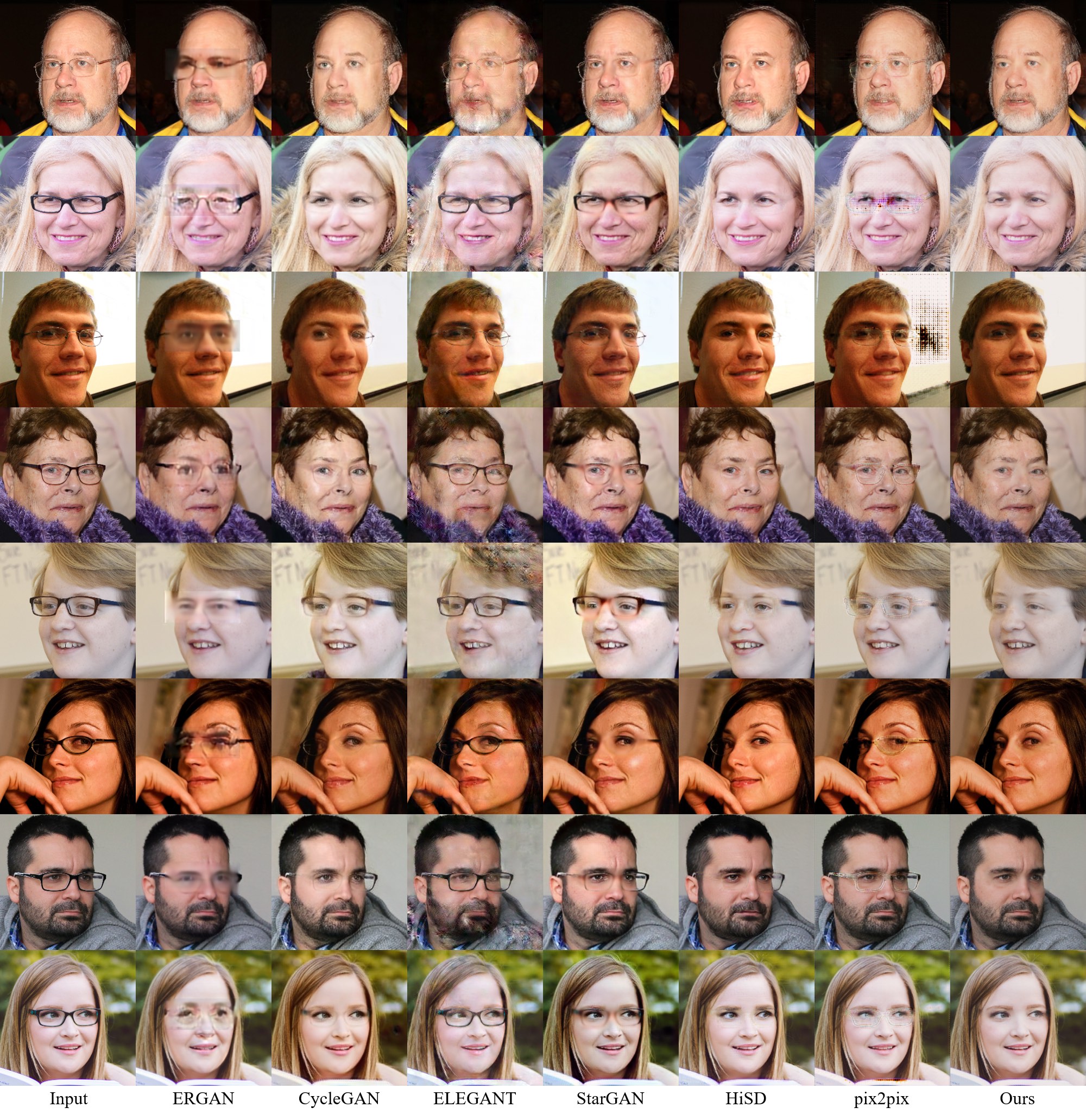}
	\caption{More results of different head poses.}
	\label{fig:headpose}
\end{figure*}

\begin{figure*}[htp]
	\centering
	\includegraphics[width=1.0\linewidth]{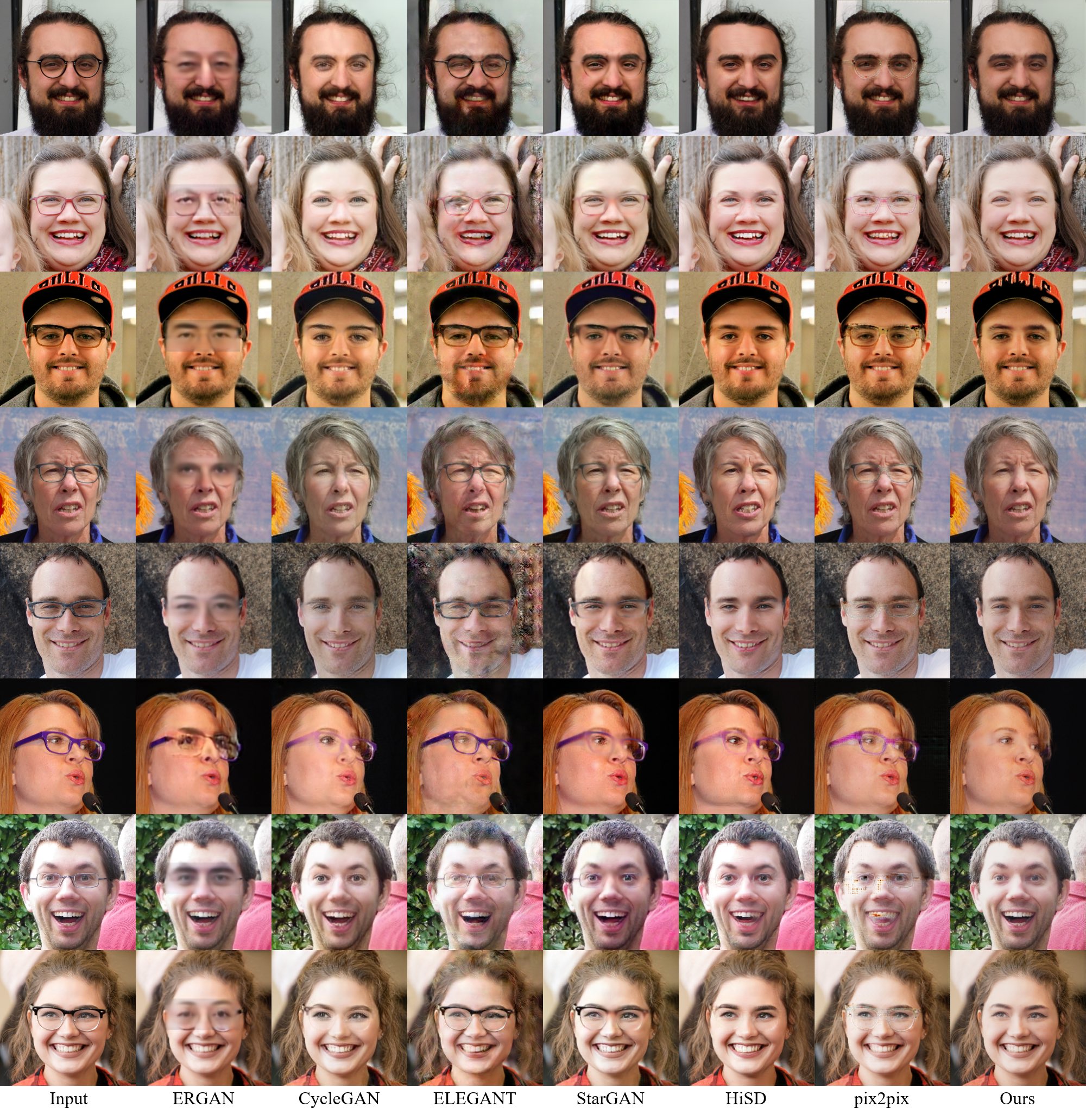}
	\caption{More results of eyeglasses with different shapes and textures.}
	\label{fig:shapetexture}
\end{figure*}

To demonstrate the robustness of our proposed method, we show more qualitative results for illuminations (\cref{fig:shadows}), head poses (\cref{fig:headpose}) and eyeglasses variation (\cref{fig:shapetexture}). 

\section{Qualitative Comparison on CelebA}
\begin{figure*}[htp]
	\centering
	\includegraphics[width=1.0\linewidth]{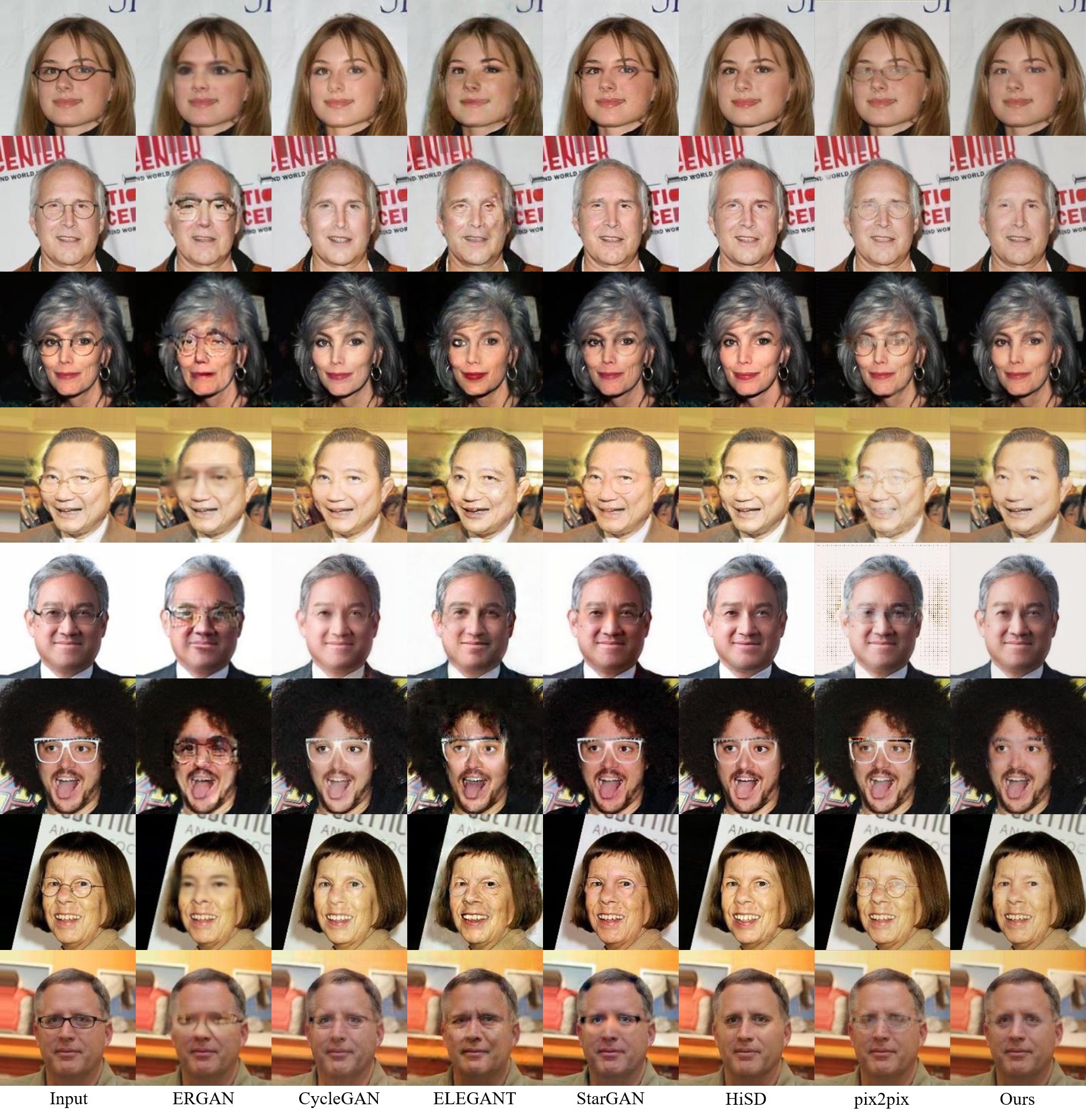}
	\caption{Qualitative comparisons on CelebA.}
	\label{fig:celeba}
\end{figure*}

Some methods in the previous comparisons, \eg, ELEGANT\cite{Xiao_2018_ECCV}, show limited performance in the task of eyeglasses removal. To validate the correct usage of their released code, we also show qualitative results on the testing set of CelebA\cite{liu2015faceattributes} using the same trained models.
Results in \cref{fig:celeba} show that most of the methods, \eg, ELEGANT, have better performance compared to their results on FFHQ\cite{karras2019style} or MeGlass\cite{guo2018face}.

\section{Eyeglasses Removal via StarGANv2}
\begin{figure}[btp]
	\centering
	\includegraphics[width=1.0\linewidth]{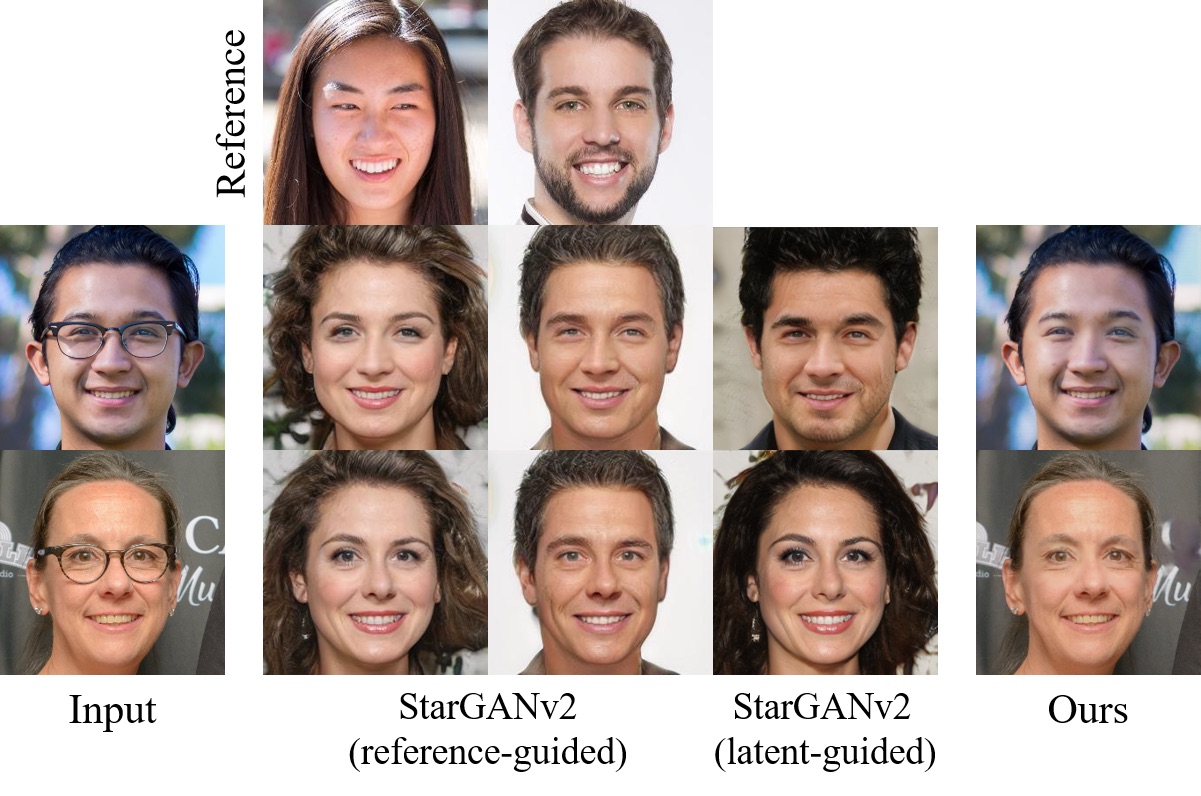}
	\caption{Comparison with StarGANv2 on the task of eyeglasses removal.}
	\label{fig:stargan2}
\end{figure}

StarGANv2\cite{choi2020starganv2} learns mixed style to transfer images of one domain to diverse images of a target domain. Although it inherits the assumption of StarGAN\cite{choi2018stargan}, the behaviour of the two networks is quite different.
In practice, we adopt StarGANv2 to the task of eyeglasses removal 
via two ways: reference-guided and latent-guided.
The results in \cref{fig:stargan2} show that StarGANv2 prefers to change other facial attributes, \eg, skin color and hairstyle, when removing the eyeglasses. Similar phenomena are also mentioned in \cite{Li_2021_CVPR}. Based on these experiments, it seems inappropriate to compare StarGANv2 on the task of eyeglasses removal. Therefore, we make comparisons with StarGAN as replace in our paper.

\section{Potential Negative Impact}
Although our method hardly changes identity, it still requires people's consent to edit their portraits for public usage. The potential negative impact might include abuse of our method without any consent. Besides, applying the technique at sufficiently large scale may cause unintended societal effects. For example, it could worsen the already widespread stigmatization of corrective eyewear. This technique should be applied in a correct way, and removing eyeglasses or not should be considered as a personal choice.

\end{document}